\documentclass[letterpaper]{article} 
\usepackage{aaai24}  
\usepackage{times}  
\usepackage{helvet}  
\usepackage{courier}  
\usepackage[hyphens]{url}  
\usepackage{graphicx} 
\usepackage{natbib}  
\usepackage{caption} 
\usepackage{algorithm}
\usepackage{algorithmic}

\usepackage{newfloat}
\usepackage{listings}
\DeclareCaptionStyle{ruled}{labelfont=normalfont,labelsep=colon,strut=off} 
\floatstyle{ruled}
\newfloat{listing}{tb}{lst}{}
\floatname{listing}{Listing}
\title{PromptMRG: Diagnosis-Driven Prompts for Medical Report Generation}

\author{Haibo Jin$^{1}$, Haoxuan Che$^{1}$, Yi Lin$^{1}$, and Hao Chen\textsuperscript{\Letter}$^{1,2}$ }

\affiliations{$^{1}$Department of Computer Science and Engineering\\
$^{2}$Department of Chemical and Biological Engineering\\
The Hong Kong University of Science and Technology, Hong Kong, China\\
\texttt{ \{hjinag,hche,jhc\}@cse.ust.hk, yi.lin@connect.ust.hk}}

\usepackage{bm}
\usepackage{amsmath}
\usepackage{amsfonts}       
\usepackage{multirow}
\usepackage{tabularx}
\usepackage{booktabs}       
\usepackage{pifont}
\newcommand{\cmark}{\ding{52}}%
\newcommand{\xmark}{\ding{56}}%
\usepackage{xcolor}         
\usepackage{graphicx}
\usepackage{subcaption}
\usepackage{marvosym}

\begin{document}

\maketitle

\begin{abstract}
Automatic medical report generation (MRG) is of great research value as it has the potential to relieve radiologists from the heavy burden of report writing. Despite recent advancements, accurate MRG remains challenging due to the need for precise clinical understanding and disease identification. Moreover, the imbalanced distribution of diseases makes the challenge even more pronounced, as rare diseases are underrepresented in training data, making their diagnostic performance unreliable. To address these challenges, we propose diagnosis-driven prompts for medical report generation (PromptMRG), a novel framework that aims to improve the diagnostic accuracy of MRG with the guidance of diagnosis-aware prompts. Specifically, PromptMRG is based on encoder-decoder architecture with an extra disease classification branch. When generating reports, the diagnostic results from the classification branch are converted into token prompts to explicitly guide the generation process. To further improve the diagnostic accuracy, we design cross-modal feature enhancement, which retrieves similar reports from the database to assist the diagnosis of a query image by leveraging the knowledge from a pre-trained CLIP. Moreover, the disease imbalanced issue is addressed by applying an adaptive logit-adjusted loss to the classification branch based on the individual learning status of each disease, which overcomes the barrier of text decoder's inability to manipulate disease distributions. Experiments on two MRG benchmarks show the effectiveness of the proposed method, where it obtains state-of-the-art clinical efficacy performance on both datasets. The code is available at https://github.com/jhb86253817/PromptMRG.
\end{abstract}

\vspace{-2mm}
\section{Introduction}
\label{sec1}

Automated analysis of medical images involves wide range of tasks, such as anomaly detection~\cite{CCY22}, disease classification~\cite{LXC22,LYC20}, lesion detection~\cite{LCZ21}, landmark detection~\cite{JCC23}, etc. Among them, medical report generation (MRG) is a task to automatically generate a free-text description of a medical image (e.g., chest X-ray), where it provides a comprehensive and contextually relevant summary of the image’s content. Due to its potential in relieving the heavy workload of radiologists, many works haven been proposed for MRG in recent years.

However, it is challenging to generate an accurate medical report as it demands a comprehensive understanding of the given image, especially the ability to identify clinical findings. For example, Figure 1(a) shows two sample predictions of a chest X-ray alongside the ground-truth (GT). While the wording of the first prediction is highly similar to the GT, its diagnosis regarding opacity and pneumonia is incorrect. In contrast, the second prediction is preferred as it successfully identifies opacity and pneumonia, albeit the different wording. Therefore, an ideal MRG system should be able to identify abnormalities accurately, then convert the findings into texts with both linguistic precision and clinical relevance.

\begin{figure}[t]
\centering
  \includegraphics[width=1\linewidth]{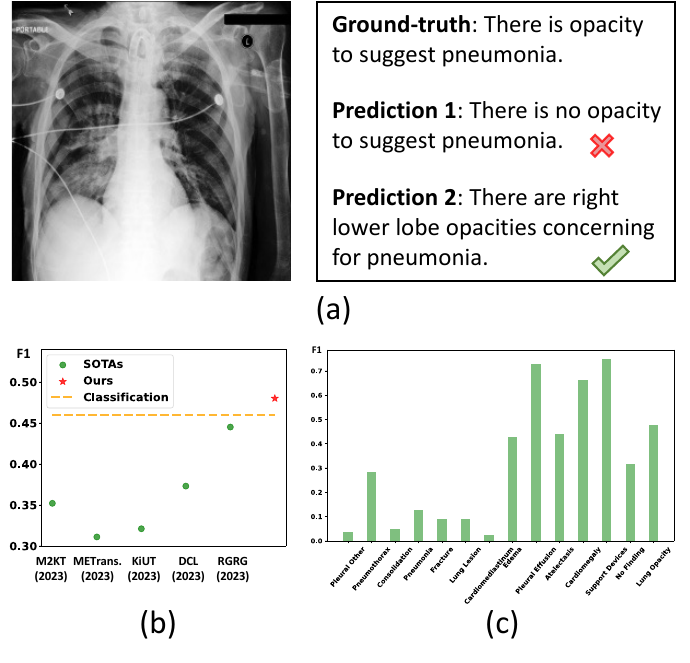}
\vspace{-7mm}
\caption{(a) Comparison of two sample predictions. (b) F1 scores of five SOTA methods published in 2023, a vanilla classification model, and our proposed model, tested on MIMIC test set. (c) F1 scores of a vanilla MRG model over different diseases on MIMIC test, and diseases are sorted in ascending order of training numbers.}
\label{fig:intro}      
\end{figure}

To obtain a MRG system with satisfactory performance, various methods have been proposed. For example, knowledge graph is an effective technique to enhance feature learning and diagnostic ability by injecting domain knowledge into the model~\cite{ZWX20,LWG21}; multi-task learning has also been widely used for obtaining better feature representations, where extra auxiliary tasks are simultaneously conducted in addition to report generation~\cite{JXX18,WHW22,YaP22}. Despite the success, state-of-the-art (SOTA) methods still lack the ability in generating diagnostically correct reports. As evidenced by our preliminary experiments shown in Figure 1(b), a vanilla disease classification model outperforms most SOTA MRG methods significantly in terms of the F1 score of clinical efficacy (CE). In MRG, CE serves as a metric for assessing the diagnostic accuracy of generated reports. Thus, the figure indicates the existing MRG methods have not fully leveraged the diagnostic information in medical images (when compared to classification models), which is an obstacle to the application of MRG. Additionally, the biased distribution of diseases leads to imbalanced CE performance (see Figure 1(c)). Yet, this issue has not been explicitly addressed in previous works, which further reduces the clinical value of current MRG models as their diagnosis on rare diseases are unreliable. 

Inspired by the above observations, we propose PromptMRG, a medical report generation framework with diagnosis-driven prompts (DDP), aiming to improve the CE performance of MRG with the guidance of diagnostic results. Specifically, 
based on the encoder-decoder architecture, PromptMRG is also equipped with a disease classification branch. When generating reports, the diagnostic results from the classification branch are converted into token prompts to explicitly guide the generation process. To further improve the diagnostic accuracy, we design cross-modal feature enhancement (CFE), which retrieves similar reports from the database to assist the diagnosis of a query image by leveraging a pre-trained CLIP model. Moreover, the disease imbalanced issue is also explicitly addressed via self-adaptive disease-balanced learning (SDL), which adaptively adjusts the optimization objectives of different diseases based on their learning status. Experiments on two MRG benchmarks show the effectiveness of the proposed method, where it obtains SOTA CE performance on both datasets. We summarize contributions as follows. 

\begin{itemize}
\item We propose a new MRG framework that utilizes a disease classification branch to guide the report generation process via token prompts, enabling the model to produce diagnostically correct reports.
\item A feature enhancement module is designed to improve the disease classification performance by leveraging the multi-modal knowledge from a pre-trained foundation model for similar records retrieval.
\item Self-adaptive disease-balanced learning is proposed to address the imbalanced learning among diseases by applying an adaptive logit-adjusted loss to the classification branch, which overcomes the barrier of text decoder's inability to manipulate disease distributions.
\item We demonstrate the superiority of PromptMRG through two popular benchmarks, where it obtains SOTA CE performance on both datasets. 
\end{itemize}

\begin{figure*}[t]
\centering
  \includegraphics[width=1\linewidth]{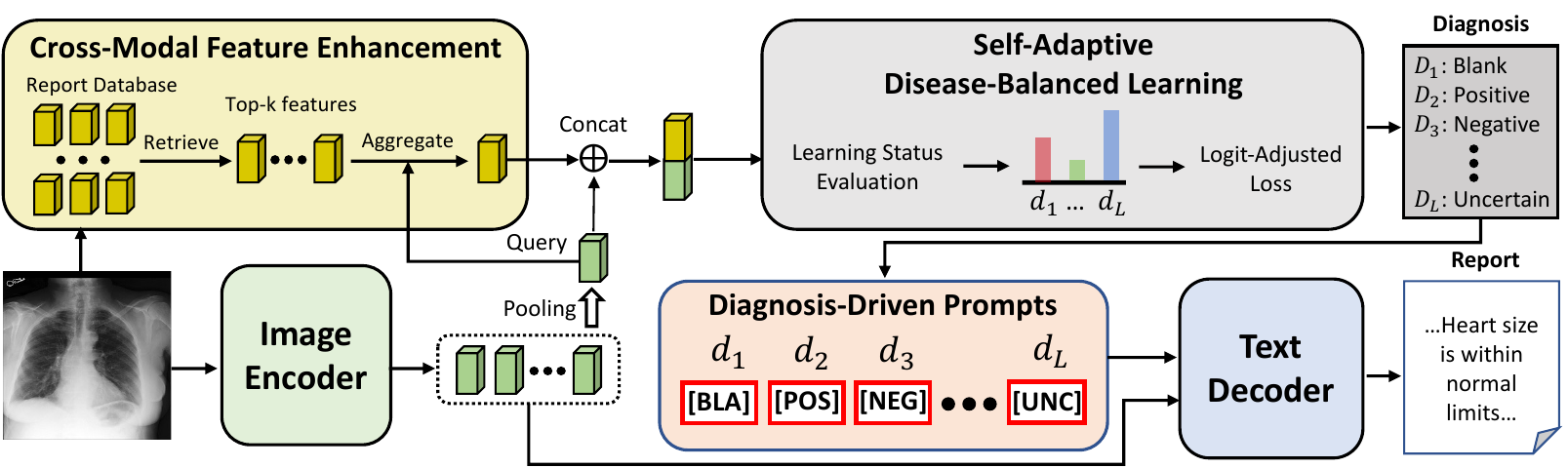}
\caption{The overall framework of PromptMRG, which mainly consists of an image encoder and a text decoder for report generation. The diagnosis-driven prompts module is proposed to guide the decoder for diagnostically correct reports. The cross-modal feature enhancement is designed to enhance the feature for disease classification via a report database. The self-adaptive disease-balanced learning is further proposed to handle the imbalanced performance among diseases.}
\label{fig:framework}      
\end{figure*}

\vspace{-3mm}
\section{Related Works}
\label{sec2}

\subsection{Medical Report Generation}
\label{sec2.1}

Most MRG models adopted the encoder-decoder architecture from image captioning~\cite{XBK15, LXP17,JLS21} due to the similarity of the two tasks. However, MRG is more challenging than image captioning because medical reports are usually much longer than captions while the clinical abnormalities in medical images are more difficult to identify than natural objects. Therefore, various methods have been proposed to tackle the above challenges. \citet{CSC20} and \citet{YWG23} proposed extra memory modules to record past similar patterns for providing informative content during the decoding process, such that the generation performance could be improved. The proposed CFE in this paper also retrieves similar records as extra information, but differently, it utilizes these information to enhance the disease classification branch rather than the generation process\cite{CSC20,YWG23}. 

Knowledge graph has been widely used to incorporate domain knowledge to assist report generation. For example, \citet{ZWX20} and \citet{LWG21} proposed to combine a pre-constructed graph to denote the relationship between diseases and organs via graph neural networks, which allows for dedicated feature learning of the abnormalities. Later, \citet{LLC23} developed a dynamic method instead of fixed ones, which dynamically updates the graph by injecting new knowledge on-the-fly. \citet{HZZ23} designed an injected knowledge distiller to fuse the knowledge from a symptom graph into the final decoding stage, which shares a similar spirit to the DDP in this paper. Nevertheless, DDP explicitly tackles the clinical efficacy issue via a different guidance mechanism (i.e., prompts), and shows much stronger performance in clinical efficacy.

Multi-task learning is another common technique to facilitate the representation learning of MRG. Among the auxiliary tasks, disease classification is the most popular one as it helps model to learn discriminative features~\cite{JXX18,WHW22,YaP22}. Similarly, weakly supervised contrastive learning was introduced by \citet{YHL21} as an auxiliary task to learn a semantically meaningful space for better report generation. Additionally, image-text matching was explored~\cite{WHW22,WZW21,YaP22} to learn an aligned image-text representations in a fine-grained manner. Despite the usage of disease classification in this work, we highlight the key difference as follows. Previous methods often treat the classification branch as a parallel task and expect it to benefit the report generation task in an implicit way through learning discriminative features. In contrast, we make use of the diagnostic results from the classification via prompts to explicitly guide the generation process. RGRG~\cite{TMK23} is the most related work to ours, which leverages object detector as a region guidance for sentence-wise generation. However, their decoder only attends to the regional visual features as most previous works do while ours attends to both visual features and diagnosis-driven prompts, where the prompts enable the decoder to explicitly leverage the diagnostic information for generating clinically correct reports. Moreover, our model can be optimized in an end-to-end manner while RGRG requires separate training of several modules and heuristic report fusion from sentence candidates.  

\vspace{-2mm}
\subsection{Prompt as Guidance}
\label{sec2.2}

Prompting is originally a technique from natural language processing for improving the generalization of language models~\cite{LYF23}.
Instead of training various tasks in supervised learning individually, prompting enables language models to unify and adapt to a wide range of tasks by modifying inputs into textual templates. Later, some works~\cite{LiL21,LAC21,LJF21} adopted this technique for efficient fine-tuning, where prompts act as trainable task-specific vectors. Due to the effectiveness and simplicity, prompt tuning was further introduced to vision~\cite{JTC22} and vision-language models~\cite{RKH21,ZYL22,TMC21,ADL22}. More recently, there are works treating prompts as a guidance for improving the performance of specific tasks. For example, \citet{QYL23} developed an automatic generation method of medical prompts to improve the knowledge transferability of pre-trained vision-language models to medical object detection. \citet{GHX22} proposed to embed domain information into prompts for unsupervised domain adaptation, so that the model learns domain-specific features according to the embedded domain information. In this paper, we propose to convert diagnostic results into prompts to guide report generation. To the best of our knowledge, this is the first work that introduces prompts to the task of MRG.

\section{Method}
\label{sec3}

In this section, we first introduce the overall framework of our model, then present the proposed modules, namely DDP, CFE, and SDL, respectively.

\vspace{-2mm}
\subsection{Framework}
\label{sec3.1}
\vspace{-1mm}

The overall architecture is shown in Figure 2. As can be seen, PromptMRG follows the mainstream encoder-decoder architecture, where the encoder $f_e$ extracts visual features of an image $I$ and the decoder $f_d$ generates report $R$ conditioned on both visual features and diagnosis-driven prompts. Formally, we denote the visual feature extraction as 
\begin{equation}
f_e(I) = \bm{X} = \{\bm{x}_1, \bm{x}_2, ..., \bm{x}_S\},
\end{equation}
where $\bm{x}_i \in \mathbb{R}^{C}$ is a feature patch, $C$ denotes the feature dimension, and $S$ is the number of patches. We denote each report as $R=\{ r_1, r_2, ..., r_T \}, r_i \in \mathbb{V}$, where each $r_i$ is a token, $T$ is the length of the report, and $\mathbb{V}$ represents the vocabulary. The process of decoding is formulated as
\begin{equation}
r_t = f_d(\bm{X}, d_1,...,d_L, r_1,...,r_{t-1}),
\end{equation}
where $r_t$ is the token to be predicted at time step $t$ and $\{ d_1,...,d_L \}$ are diagnosis-driven prompts (see next subsection). Language modeling loss is used as the primary loss:
\begin{equation}
\mathcal{L}_{\text{LM}} = -\sum_{t=1}^T \log p(r_t|r_1,...,r_{t-1},X, d_1,...,d_L).
\end{equation}

\vspace{-3mm}
\subsection{Diagnosis-Driven Prompts}
\label{sec3.2}
\vspace{-1mm}
 
Generating texts that are consistent with the diagnostic results is essential to the task of MRG. This is because the report of a medical image should not only provide a comprehensive summary, but also reflect the clinical significance. If the generated report is not diagnostically accurate, it may give a wrong conclusion of an exam, which can lead to serious consequences. However, we have observed that existing models have difficulty in generating reports with satisfactory clinical efficacy. Specifically, we trained a vanilla disease classification model on MIMIC training set and compared it with SOTA MRG methods on MIMIC test set in terms of F1 score. From the result shown in Figure 1(b), we see that the classification model outperforms most SOTA methods by a large margin, indicating that the existing MRG models still lack the ability in generating diagnostically accurate reports. 

Motivated by the above observation, we propose diagnosis-driven prompts (DDP), which act as a guidance of the text decoder by conveying the diagnostic results from a disease classification branch. The branch takes as inputs the average pooled visual features with cross-modal enhancement (see next subsection), and outputs the classification results via $L$ classification heads, where $L$ is the number of diseases. Each classification head conducts a 4-class classification task via a fully connected layer, namely 'Blank', 'Positive', 'Negative', and 'Uncertain'. The classification labels can be obtained with CheXbert~\cite{SJR20} by converting reports into 14 predefined disease labels and the training can be done with standard cross-entropy loss $\mathcal{L}_{\text{CE}}$.  

During inference, the classification results are converted into token prompts, where each token prompt corresponds to one disease. To achieve this, we add four new tokens, \texttt{[BLA]}, \texttt{[POS]}, \texttt{[NEG]}, and \texttt{[UNC]}, to the vocabulary to represent the four classification classes, respectively. In this way, the decoder can explicitly refer to these prompts for generating reports with better clinical efficacy. We also give quantitative results to compare different types of prompts and qualitative examples to show how the prompts guide the generation process via attention weights (see Experiments).  

\begin{figure}[t]
\centering
  \includegraphics[width=1\linewidth]{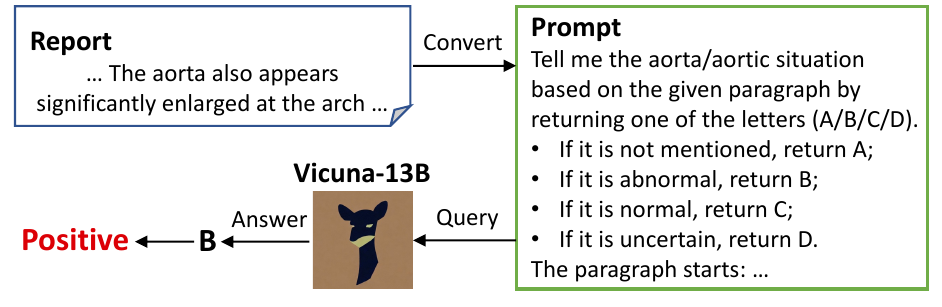}
\vspace{-7mm}
\caption{An example prompt we used to query the label of Aorta from Vicuna-13B.}
\label{fig:llm}      
\end{figure}

\paragraph{Auxiliary disease labeling with LLMs.}
We found that the reports of our training data cover more abnormalities than the predefined 14 diseases, and these extra disease information can also be beneficial to accurate diagnosis. Thus, we leverage the ability of large language models (LLMs) to obtain labels of four auxiliary abnormalities, including 1) Aorta, 2) Bone/Spine, 3) Hemidiaphragm, and 4) Lung Volume. Specifically, we used Vicuna-13B~\cite{ZCS23} as our labeling assistant, and we present it reports with disease related prompts for querying the label of target diseases. Figure 3 shows an example prompt that we used for auxiliary disease labeling.

\vspace{-2mm}
\subsection{Cross-Modal Feature Enhancement}
\label{sec3.3}

Solely based on medical images for diagnosis can be suboptimal, as radiologists can usually access extra documents for references, such as patient information and diagnostic database. Inspired by this, in addition to the visual features, we also resort to the report database of training data to obtain more robust features for disease classification. To this end, we propose cross-modal feature enhancement (CFE), and its architecture is shown in Figure 4. In CFE, we first leverage a CLIP model pre-trained on MIMIC training set~\cite{EKK21} to perform cross-modal retrieval, which gives us top-$k$ report features of a given image $I$. Then these features are aggregated into an embedding via a dynamic aggregation (DA) module, which is further concatenated with the visual feature for disease classification. Formally, we have
\begin{equation}
\bm{x}^E = \text{DA}(\bm{X}^{\prime}, \bm{x}^V) \oplus \bm{x}^V,
\end{equation}
where $\bm{X}^{\prime}=\{ \bm{x}_1^{\prime},...,\bm{x}_k^{\prime} \}$ is the retrieved top-$k$ report features, $\bm{x}^V$ is the average pooled visual feature, and $\oplus$ represents concatenation. To dynamically extract report features with respect to the visual feature $\bm{x}^V$, we implement DA as Transformer attention modules~\cite{VSP17}. $\bm{X}^{\prime}$ first goes through a self-attention layer, then the output is used as the key and value of a cross-attention layer while $\bm{x}^V$ is the query. Thus, DA can be formulated as 
\begin{equation}
\text{DA}(\bm{X}^{\prime}, \bm{x}^V) = \text{Cross-Attn}(\text{Self-Attn}(\bm{X}^{\prime}), \bm{x}^V).
\end{equation}
Note that the DA module is trainable while the CLIP is frozen during training. We argue that the philosophy of CFE is similar to the workflow of radiologists. In clinical practice, when radiologists are uncertain about the diagnosis of a new medical image, they may refer to past records for comparison and verification so that the results are more reliable. 

\begin{figure}[t]
\centering
  \includegraphics[width=1\linewidth]{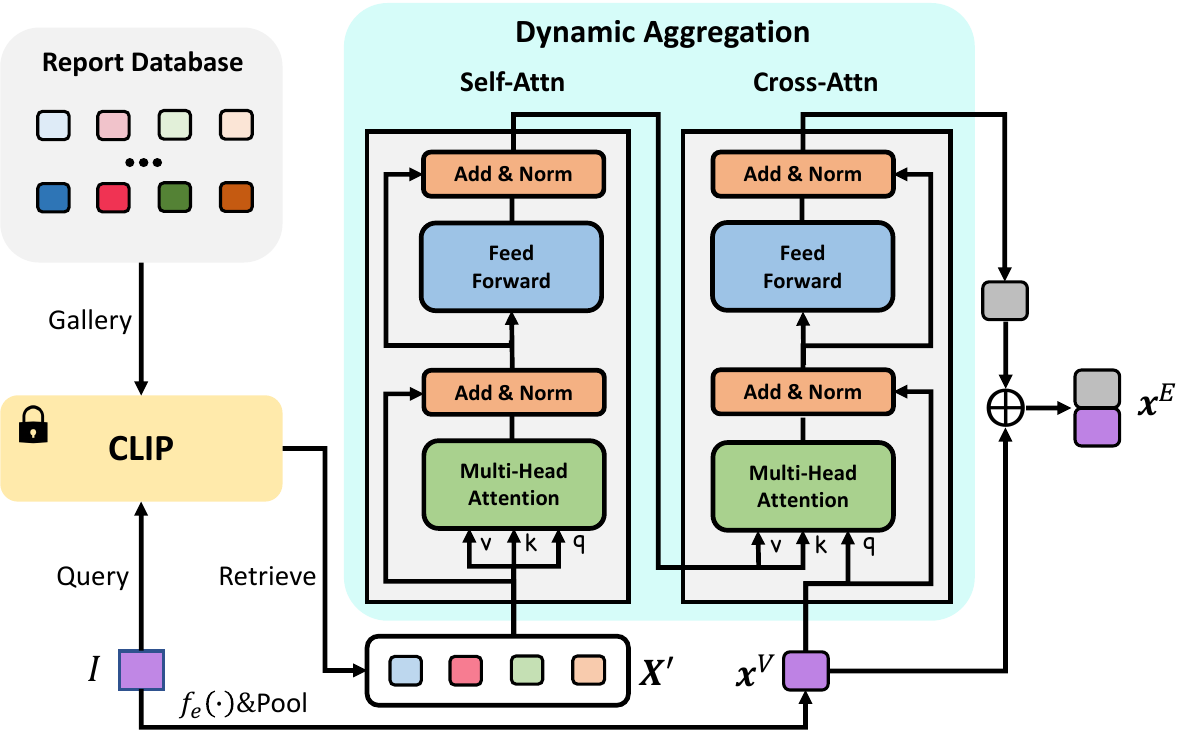}
\vspace{-5mm}
\caption{Architecture of cross-modal feature enhancement.}
\label{fig:cfe}      
\end{figure}

\begin{table*}[t]
\centering
\footnotesize
\newcolumntype{C}{>{\centering\arraybackslash}X}%
\begin{tabularx}{\linewidth}{lllCCCCCCC}
\toprule	 
\multirow{2}{*}{\textbf{Dataset}} & \multirow{2}{*}{\textbf{Model}} & \multirow{2}{*}{\textbf{Year}} & \multicolumn{3}{c}{\textbf{CE Metrics}} & \multicolumn{4}{c}{\textbf{NLG Metrics}}\\ \cmidrule(r){4-6} \cmidrule(r){7-10}
 & & &Precision & Recall & F1 & BLEU-1 & BLEU-4 & METEOR & ROUGE-L \\
\midrule
\multirow{11}{*}{\textbf{MIMIC}} & R2Gen & 2020 & 0.333 & 0.273 & 0.276 & 0.353 & 0.103 & 0.142 & 0.277 \\
& M2TR & 2021 & 0.240 & 0.428 & 0.308 & 0.378 & 0.107 & 0.145 & 0.272 \\
& MKSG & 2022 & 0.458 & 0.348 & 0.371 & 0.363 & 0.115 & - & 0.284 \\
& CliBert & 2022 & 0.397 & 0.435 & 0.415 & 0.383 & 0.106 & 0.144 & 0.275 \\
& CVT2Dis.$^*$ & 2022 & 0.356 & 0.412 & 0.384 & 0.392 & 0.124 & 0.153 & 0.285 \\
& M2KT & 2023 & 0.420 & 0.339 & 0.352 & 0.386 & 0.111 & - & 0.274 \\
& METrans. & 2023 & 0.364 & 0.309 & 0.311 & 0.386 & 0.124 & 0.152 & \textbf{0.291} \\
& KiUT & 2023 & 0.371 & 0.318 & 0.321 & 0.393 & 0.113 & 0.160 & 0.285 \\
& DCL & 2023 & 0.471 & 0.352 & 0.373 & - & 0.109 & 0.150 & 0.284 \\
& RGRG$^{*\#}$ & 2023 & 0.461 & 0.475 & 0.447 & 0.373 & \textbf{0.126} & \textbf{0.168} & 0.264 \\
\cmidrule(r){2-10}
& \textbf{Ours} & - & \textbf{0.501} & \textbf{0.509} & \textbf{0.476} & \textbf{0.398} & 0.112 & 0.157 & 0.268 \\
\midrule
\multirow{6}{*}{\textbf{IU X-Ray}} & R2Gen$^{\dagger}$ & 2020 & 0.141 & 0.136 & 0.136 & 0.325 & 0.059 & 0.131 & 0.253 \\
& CVT2Dis.$^{*\dagger}$ & 2022 & 0.174 & 0.172 & 0.168 & 0.383 & 0.082 & 0.147 & 0.277 \\
& M2KT$^{\dagger}$ & 2023 & 0.153 & 0.145 & 0.145 & 0.371 & 0.078 & 0.153 & 0.261 \\
& DCL$^{\dagger}$ & 2023 & 0.168 & 0.167 & 0.162 & 0.354 & 0.074 & 0.152 & 0.267 \\
& RGRG$^{*\dagger}$ & 2023 & 0.183 & 0.187 & 0.180 & 0.266 & 0.063 & 0.146 & 0.180 \\
\cmidrule(r){2-10}
& \textbf{Ours} & - & \textbf{0.213} & \textbf{0.229} & \textbf{0.211} & \textbf{0.401} & \textbf{0.098} & \textbf{0.160} & \textbf{0.281} \\
\bottomrule
\end{tabularx}
\vspace{-2mm}
\caption{\small Comparison with SOTA MRG methods on MIMIC-CXR and IU X-Ray. $*$ indicates the used image size is larger than 224. $\dagger$ indicates the performance evaluated by us. $\#$ indicates the usage of a different data split. The best results are in \textbf{bold}.}
\label{tab:results_sota}
\vspace{-2mm}
\end{table*}

\vspace{-2mm}
\subsection{Self-Adaptive Disease-Balanced Learning}
\label{sec3.4}

Due to the biased nature of diseases, some abnormalities are common to see in the reports while some are quite rare. Such an imbalanced distribution would lead to imbalanced learning of diseases, where the common diseases can be well identified but the rare ones are of poor performance. This also applies to the largest MRG dataset MIMIC~\cite{CSC20,NDK22}. To verify it, we trained a vanilla MRG model (i.e., our baseline) on MIMIC, and plot the F1 score of 14 diseases individually. As can be seen from Figure 1(c),  the performance gap between different diseases can be quite large. This imbalanced problem makes the diagnosis of rare diseases unreliable, thus considerably affecting the clinical application of MRG. Although this problem has been noted by some prior works~\cite{CSC20,NDK22}, no solution has been proposed to address this issue explicitly. It may be due to the insensitivity of the text decoder to diseases, as it generates words solely based on likelihoods without distinguishing between different diseases. Thus, it is not straightforward to manipulate the distribution of diseases through the decoder. 

In this paper, we aim to address the imbalanced issue via the classification branch as it is directly related to the learning of different diseases. When the imbalanced issue is handled within the classification branch, the disease-balanced results could then be used to guide the report generation via the proposed prompts. To this end, we propose self-adaptive disease-balanced learning (SDL), an algorithm that adaptively adjusts the learning objectives of different diseases based on their learning status. To balance the learning between diseases, we introduce the logit-adjusted loss~\cite{MJR20}, which encourages rare diseases to learn more by decreasing their logits during optimization. For a given disease $D$, its logit-adjusted loss against label $P$ (i.e., Positive) is formulated as
\begin{equation}
\begin{split}
& \ell_D(y=P, f(\bm{x}^E)) \\
& = -\log \frac{e^{f_y(\bm{x}^E)+ \log \pi_D}}{\sum_{y^{\prime} \neq P} e^{f_{y^{\prime}}(\bm{x}^E)} + (e^{f_y(\bm{x}^E)+ \log \pi_D})},
\end{split}
\end{equation}
where $f_y(\bm{x}^E)$ is the logit of class $y$, and $\pi_D$ is the class distribution of disease $D$. The loss against non-positive labels remains the same as $\mathcal{L}_{\text{CE}}$. 

However, the fixed class distribution~\cite{MJR20} used for logit adjustment cannot reflect the learning dynamics of diseases because these diseases not only have varied distributions, but are also with different learning difficulties. This issue is also depicted in Figure 1(c), where diseases are arranged in ascending order of training numbers, and their performance does not necessarily improve as the training number increases. Inspired by \citet{ZWH21}, we propose to utilize prediction scores as an assessment of the learning status for different diseases: a large score indicates a well-learned disease while a small one implies that the disease is not sufficiently learned. Thus, the class distribution $\pi$ can be initialized with the statistics from training data, then adaptively updated with the following formula:  
\begin{equation}
\pi(j) = [s_1^j, s_2^j, ..., s_L^j],
\end{equation}
where $s_i^j$ is the average prediction score of the $i$-th disease on the validation set at the $j$-th epoch. We denote the loss of SDL as $\mathcal{L}_{\text{SDL}}$, and the total training loss of our model is
\begin{equation}
\mathcal{L} = \mathcal{L}_{\text{LM}} + \lambda \mathcal{L}_{\text{SDL}},
\end{equation}
where $\lambda$ is the balancing coefficient.

\section{Experiments}
\label{sec4}

\subsection{Datasets}
\vspace{-1mm}

\textbf{1) MIMIC-CXR}~\cite{JPG19} is the largest MRG dataset with chest X-ray images and paired reports. We follow the official split and the preprocessing from \citet{CSC20}, where the processed dataset has 270,790, 2,130, and 3,858 samples for training, validation, and test, respectively. \textbf{2) IU X-Ray}~\cite{DKR16} is also widely used for MRG evaluation, which contains 2,955 samples in total after preprocessing. We found the test split from \citet{CSC20} is not suitable for disease-aware evaluation as some diseases only have a couple of positive samples. Therefore, we use the model trained on MIMIC-CXR training set to directly perform the evaluation on the whole set of IU X-Ray. See IU X-Ray Dataset in the Appendix for more details.

\begin{figure*}[t]
\centering
  \includegraphics[width=1\linewidth]{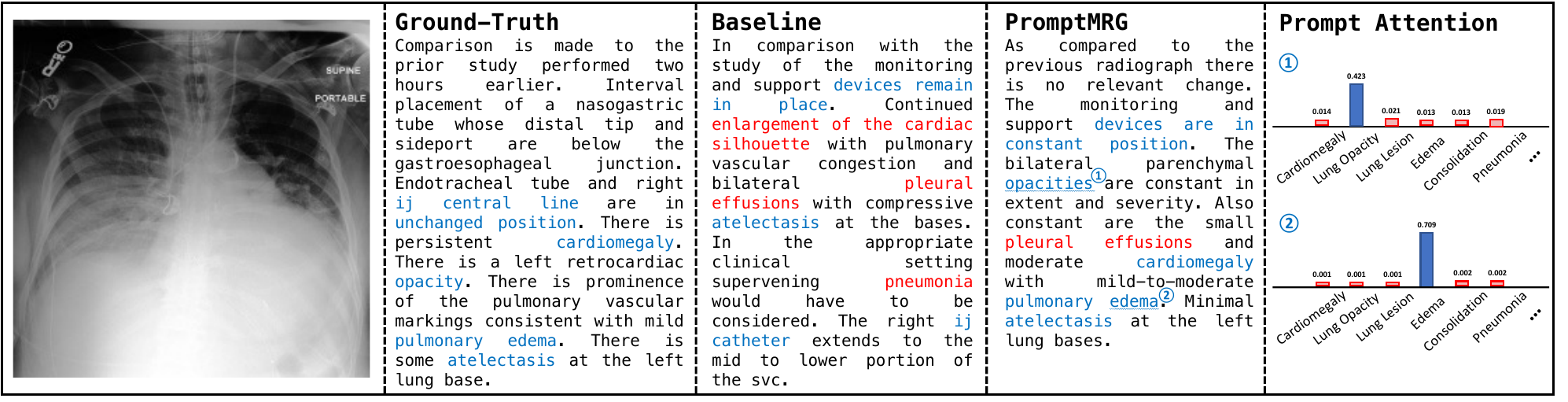}
\vspace{-5mm}
\caption{Qualitative examples of the baseline and the proposed method. Blue font indicates consistent content with the ground-truth while red font indicates incorrect content. See more examples in Appendix.}
\label{fig:qual}      
\end{figure*}

\begin{table*}[t]
\centering
\footnotesize
\newcolumntype{C}{>{\centering\arraybackslash}X}%
\begin{tabularx}{\linewidth}{llllCCCCCCC}
\toprule	 
\multirow{2}{*}{DDP} & \multirow{2}{*}{ADL} & \multirow{2}{*}{CFE} & \multirow{2}{*}{SDL} & \multicolumn{3}{c}{\textbf{CE Metrics}} & \multicolumn{4}{c}{\textbf{NLG Metrics}}\\ \cmidrule(r){5-7} \cmidrule(r){8-11}
 & & & &Precision & Recall & F1 & BLEU-1 & BLEU-4 & METEOR & ROUGE-L \\
\midrule
\xmark & \xmark & \xmark & \xmark & 0.430 & 0.368 & 0.370 & 0.397 & \textbf{0.116} & \textbf{0.158} & \textbf{0.272} \\
\cmark & \xmark & \xmark & \xmark & 0.487 & 0.461 & 0.444 & 0.391 & 0.106 & 0.151 & 0.261 \\
\cmark & \cmark & \xmark & \xmark & 0.496 & 0.466 & 0.451 & 0.393 & 0.106 & 0.152 & 0.261 \\
\cmark & \cmark & \cmark & \xmark & \textbf{0.514} & 0.475 & 0.464 & 0.394 & 0.108 & 0.152 & 0.263 \\
\cmark & \cmark & \xmark & \cmark & 0.489 & 0.500 & 0.468 & 0.397 & 0.111 & 0.155 & 0.264 \\
\cmark & \cmark & \cmark & \cmark & 0.501 & \textbf{0.509} & \textbf{0.476} & \textbf{0.398} & 0.112 & 0.157 & 0.268 \\
\bottomrule
\end{tabularx}
\vspace{-2mm}
\caption{\small Ablation study of each module on MIMIC test set.}
\label{tab:results_ablation}
\vspace{-3mm}
\end{table*}

\subsection{Evaluation Metrics}
\vspace{-1mm}

We evaluate model performance with both natural language generation (NLG) metrics and CE metrics. The NLG metrics include BLEU~\cite{PRW02}, METEOR~\cite{DeL11}, and ROUGE-L~\cite{Lin04}. Following \citet{NDK22}, the CE metrics include precision, recall, and F1, which are evaluated by converting reports into 14 disease classification labels using CheXbert~\cite{SJR20}.

\subsection{Implementation details}
\vspace{-1mm}

We use ImageNet pre-trained ResNet-101~\cite{HZR16} as the encoder and Bert-base~\cite{DCL19} as the decoder. The coefficient $\lambda$ of $\mathcal{L}_{\text{SDL}}$ is set to 4 and $k=21$ is used for CFE. The sensitivity of $\lambda$ and $k$ is analyzed in Appendix. AdamW~\cite{LoH17} is used as the optimizer with a weight decay of 0.05. The initial learning rate is set to 5e-5 with a cosine learning rate schedule. The number of training epochs is 10, batch size is 16, and image size is 224. The model was implemented with PyTorch 2.0 and trained with one RTX 3090 GPU for about 24 hours. 

\subsection{Results}
\vspace{-1mm}

We compare our model with SOTA methods, including R2Gen~\cite{CSC20}, M2TR~\cite{NGF21}, MKSG~\cite{YWG22}, CliBert~\cite{YaP22}, CVT2Dis.~\cite{NDK22}, M2KT~\cite{YWG23}, METrans.~\cite{WLW23}, KiUT~\cite{HZZ23}, DCL~\cite{LLC23}, and RGRG~\cite{TMK23}. Table 1 shows the results on both MIMIC-CXR and IU X-Ray. We can see from the table that the proposed method achieves SOTA performance on the three CE metrics on both datasets and outperforms most existing methods by a large margin. For example, our method obtains 0.476 F1 score on MIMIC, which shows a 10\% absolute improvement over a recent work DCL (0.373 F1) and 15\% over KiUT (0.321 F1). Even compared with the best existing method RGRG (0.447 F1), the absolute improvement is 2.9\%. Similarly, on IU X-Ray, PromptMRG outperforms other methods considerably, where its absolute improvement over the best existing method RGRG is 3.1\%. As for NLG metrics, our method is also competitive. For example, PromptMRG achieves the best results on all the NLG metrics of IU and the best BLEU-1 on MIMIC. However, the performance of our method on the other NLG metrics of MIMIC is not as good as that of IU. We suspect it is caused by DDP, as we found that model without DDP tends to generate more frequent phrases from the training data than the one with DDP, especially for long reports (MIMIC is about two times longer than IU on average). As a result, the former will achieve better NLG performance than the latter, as NLG metrics evaluate language consistency between predictions and references by utilizing word matching. See Ablation Study below and N-gram Statistics in the Appendix for more details.

\subsection{Model Analysis}
\vspace{-1mm}

\paragraph{Ablation study.}
To verify the effectiveness of each module, we do ablation study on MIMIC test set, which is shown in Table 2. Firstly, we see that the baseline only achieves fair CE performance (e.g., a F1 score of 0.370). By adding DDP, the F1 score increases significantly to 0.444, indicating the effectiveness of DDP in producing diagnostically correct reports. Adding auxiliary disease labeling (ADL) also benefits CE results, improves F1 score to 0.451. When CFE and SDL are further applied, the F1 score improves to 0.464 and 0.468, respectively. Finally, utilizing all the modules yields the best F1 score of 0.476. We notice that ADL and CFE improve both precision and recall while SDL is more helpful for recall. We believe this is related to the mechanism of SDL, which encourages the learning of less-learned diseases on recalling positive cases. On the other hand, the usage of DDP degrades the NLG performance by about 0.8\% on average. Through an examination of the N-gram statistics of the generated reports (see Appendix), we have noted that the baseline is more likely to repeat the frequent phrases from the training set in comparison to the model that utilizes DDP. We suspect it is because DDP provides extra diagnostic information during the generation process, making the generated texts more diverse and less susceptible to high-frequency phrases. Consequently, the NLG performance would exhibit a decline with DDP, as a result of fewer matched phrases. When CFE and SDL are further added, the NLG performance increases, but still lags behind the baseline (except for BLEU-1). Overall, the proposed modules substantially improve the CE performance at the expense of slightly degraded NLG performance.   

\begin{table}[t]
  \footnotesize
  \centering
  \begin{tabular}{lcccccc}
    \toprule
    Prompt & B-1  & B-4 & Prec. & Rec.  & F1  & F1$\Delta$    \\
    \midrule
    None      & 0.397 & 0.116  & 0.430 & 0.368 & 0.370 & - \\ 
    Token     & 0.393  & 0.106  & 0.496 & 0.466 & 0.451 & \textbf{+8.1\%}      \\
    Text      & 0.385  & 0.103 & 0.490 & 0.461 & 0.446 & +7.6\%      \\
    Feature   & 0.393  & 0.111 & 0.418 & 0.352 & 0.357 & -1.3\%       \\
    Embed.    & 0.384  & 0.098 & 0.471 & 0.454 & 0.434 & +6.4\%         \\

    \bottomrule
  \end{tabular}
  \vspace{-2mm}
  \caption{Comparison of different prompt types.}
  \label{tab:prompt_type}
  \vspace{-4mm}
\end{table}

\vspace{-1mm}
\paragraph{Qualitative results.}
We show a qualitative example to demonstrate the superiority of PromptMRG over the baseline, which is shown in Figure 5. The blue fonts indicate the consistent diagnosis with the GT and the red ones imply incorrect results. As we can see, PromptMRG covers most key descriptions in the GT. For example, it correctly predicts the positive cases of cardiomegaly, opacity, edema, atelectasis, and the unchanged position of support devices, with only one wrong prediction on pleural effusion. In contrast, despite the professional wording, the baseline is unable to accurately predict the disease information. For example, it gives false positive cases for cardiomediastinum, pleural effusion, and pneumonia, while missing cardiomegaly and opacity. To understand how DDP boosts the diagnostic accuracy, we visualize the attention weights of the prompts, which is shown on the right of the figure. We can see that when predicting certain words, the attention on the relevant disease token is much larger than the remaining tokens, indicating that the token prompts indeed convey useful diagnostic information to the decoder during the generation process.  

\vspace{-1mm}
\paragraph{Type of prompts.}
For DDP, in addition to the proposed token prompt, we also explored other types of prompts for comparison. Table 3 gives the comparison of these prompts. \textbf{1) Text prompt} directly converts the classification results into texts, where a positive result of pneumonia can be represented as "Pneumonia: Positive;". We can see that the performance of text prompt is satisfactory, albeit slightly worse than that of the token prompt across all metrics. \textbf{2) Feature prompt} uses the average pooled feature right before the classification branch as prompt, which is supposed to be discriminative. However, our results show that feature prompt obtains even worse CE metrcis (e.g., 1.3\% reduction on F1) than baseline, implying that the decoder cannot extract diagnostic information from feature prompt.  \textbf{3) Embed. prompt} can be seen as an explicit version of feature prompt, which explicitly encodes the classification results as numbers into an embedding. For example, the $i$-th dimension is denoted as 1 if disease $i$ is not mentioned, denoted as 2 if it is positive, and so on. Since there are more dimensions than the number of diseases in an embedding, we simply pad the remaining dimensions with zeros. Despite being inferior to the token and text prompts, the embed. prompt considerably improves the CE performance (e.g., 6.4\% increase on F1), indicating that it is essential to represent the diagnostic information explicitly for an effective prompt guidance.

\begin{figure}[t]
\centering
  \includegraphics[width=1\linewidth]{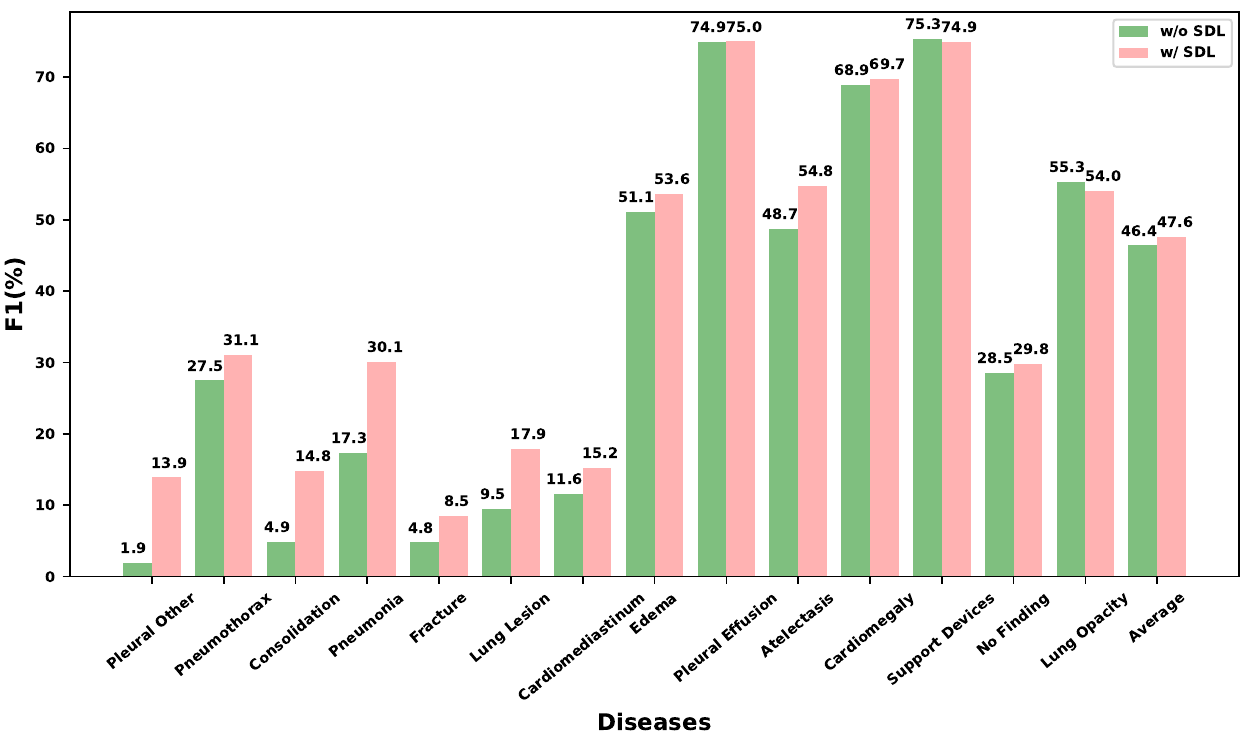}
\vspace{-6mm}
\caption{Comparison of the F1 score (\%) over the diseases between the method with and without SDL. The diseases are sorted in ascending order of their training numbers.}
\label{fig:balance}      
\vspace{-3mm}
\end{figure}

\vspace{-1mm}
\paragraph{Disease balance.}
 To assess the efficacy of SDL, we plot individual F1 scores for all diseases to compare the method with and without SDL. Figure 6 presents the results using histograms, where the diseases are sorted in ascending order of their training numbers. Firstly, we see that the performance of rare diseases have been largely improved with the help of SDL, and the average absolute improvement across seven rare diseases is approximately 8\%. Some even show improvements over 12\%, such as pleural other and pneumonia. For common diseases, some are also improved (e.g., edema and atelectasis) while some are slightly degraded (e.g., lung opacity). Despite the large improvement over rare diseases, the increase of average F1 score is only 1.2\%. It indicates that the currently adopted example-based CE metrics do not reflect the balancedness of diseases, where more frequent diseases have larger weights. Thus, while there has been a substantial increase in the performance of rare diseases, the average F1 score has only experienced a marginal improvement. To address it, macro-averaged CE metrics could be computed to reflect disease balancedness (see Appendix for Macro-Averaged Results). 
 
\vspace{-1mm}
\section{Conclusion and Future Work}
\label{sec5}

In this work, we proposed a MRG framework to tackle the problem of unsatisfactory CE, where the diagnostic results from a disease classification branch are converted into prompts to guide the report generation. The CFE module was proposed to further improve the diagnostic accuracy by enhancing the feature with cross-modal retrieval and dynamic aggregation. Furthermore, SDL was developed to alleviate the imbalanced learning among diseases by individually adjusting the learning objectives of each disease based on its unique learning status. Experiments on two datasets demonstrate the superiority of our method, especially its advantage in generating diagnostically correct reports, bridging the gap between current MRG models and the clinical demands in practice.

While the experiments were based on chest X-rays, our method is potentially applicable to other modalities. However, several issues need to be addressed. Firstly, disease labels are required for training the classification branch. Luckily, most MRG datasets contain such labels. In case no labels are provided, unsupervised clustering algorithms can be used to group reports into meaningful clusters. Secondly, the CLIP in CFE requires a sufficient amount of image-report pairs for domain adaptation, which can be challenging for domains with limited data. FFA-IR~\cite{LCL21} is an ideal dataset to validate our method on fundus images due to its large scale and disease diversity, which we leave for future work. Moreover, we would leverage richer and fine-grained information for the prompting, which could be beneficial to both linguistic precision and diagnostic accuracy.

\section{Acknowledgments}
This work was supported by the Pneumoconiosis Compensation Fund Board, HKSARS (Project No. PCFB22EG01) and Hong Kong Innovation and Technology Fund (Project No. ITS/028/21FP).

\bibliography{aaai24}

\clearpage
\appendix

\section{Appendix}
\setcounter{table}{0}
\setcounter{figure}{0}

\subsection{IU X-Ray Dataset}
\label{sec6}

\citet{CSC20} partitioned the entire IU X-Ray into train/validation/test set by 7:1:2, which is often adopted by later works~\cite{NDK22,LLC23,YWG23} for evaluation. However, we note that the test set from \citet{CSC20} is not suitable for disease-aware evaluation as some diseases, such as consolidation and pneumonia, only have two positive samples (see Table 1). Therefore, we propose to use the entire IU dataset for evaluation to ease the problem of insufficient positive diseases. In previous setting~\cite{CSC20}, each sample contains two images (i.e., frontal and lateral view) and they are often treated as one sample via concatenation. In our case, to make it compatible with the MIMIC trained model, we treat the two associated images separately, which doubles the number of samples (from 2,955 to 5,910). However, we found that the normal images (i.e., "No Finding") occupie 36\% of the test set, which makes the test biased and less informative in evaluating a model's diagnostic capability. For example, the normal images in MIMIC test only account for 8\% of the total. Thus, we randomly removed 80\% of the normal images and finally retained 414 of them. By doing this, the percentage of normal images in IU is 10\%, which is comparable to that of MIMIC. Finally, the total number of our IU test set is 4,168, and its disease-related count can found in Table 1.

\subsection{Hyperparameter Analysis}
\label{sec7}

We perform hyperparameter analysis to see the sensitivity of the two hyperparameters $\lambda$ and $k$. $\lambda$ is the balancing coefficient of loss $\mathcal{L}_{\text{SDL}}$ and $k$ represents the number of retrieved reports in the CFE module. Figure 1(a) and 1(b) show the results on MIMIC test set of $\lambda$ and $k$, respectively. In Figure 1(a), $\lambda$ is analyzed with values ranging from 0.2 to 12 in terms of F1 and BLEU-4 scores. Overall, the performance remains stable across a wide range of $\lambda$, as the fluctuations of F1 and BLEU-4 are within 1.2\% and 0.5\%, respectively. $\lambda=4$ gives the best F1 and BLEU-4 scores, which is the value we used in the experiments. When $\lambda$ is too small, the training of the classification branch might be affected, thus leading to worse F1 and BLEU-4 performance. On the other hand, a larger $\lambda$ may benefit the training of the classification branch, but the language modeling task is not well learned, making it difficult to fuse the diagnostic information from the classification branch. In Figure 1(b), $k$ is analyzed with values ranging from 1 to 81 in terms of F1 and BLEU-4 scores. Similarly, the performance also remains stable across a wide range of $k$. The optimal performance is reached at around $k=20$. When $k$ decreases, both F1 and BLEU-4 scores decrease quickly, as there is less cross-modal information for aggregation, making the model less robust; when $k$ increases, the F1 and BLEU-4 scores slowly decrease, indicating that too much cross-modal information would bring noise and make the aggregation module less effective.

\begin{table}[t]
  \footnotesize
  \centering
  \begin{tabular}{lccc}
    \toprule
    Disease & MIMIC Test & IU Test(Chen) & IU Test(Ours)      \\
    \midrule
    Enlarged Cardio. & 730 & 6 & 112 \\ 
	Cardiomegaly & 1,271 & 63 & 676 \\
	Lung Opacity & 1,392 & 59 & 784 \\
	Lung Lesion & 199 & 15 & 222 \\
	Edema & 563 & 6 & 42 \\
	Consolidation & 176 & 2 & 50 \\
	Pneumonia & 165 & 2 & 32 \\
	Atelectasis & 841 & 21 & 258 \\
	Pneumothorax & 75 & 2 & 22 \\
	Pleural Effusion & 1,056 & 12 & 142 \\
	Pleural Other & 122 & 4 & 68 \\
	Fracture & 148 & 13 & 182 \\
	Support Devices & 1,345 & 20 & 228 \\
	No Finding & 323 & 295 & 414 \\
	\midrule
	Total & 3,858 & 590 & 4,168\\
    \bottomrule
  \end{tabular}
  \vspace{-2mm}
  \caption{The positive disease count of MIMIC test~\cite{NDK22}, IU test from \citet{CSC20}, and IU test proposed by us.}
  \label{tab:iu_dataset}
\end{table}

\begin{figure}[H]
  \centering
  \begin{subfigure}[b]{1\linewidth}
	\includegraphics[width=\textwidth]{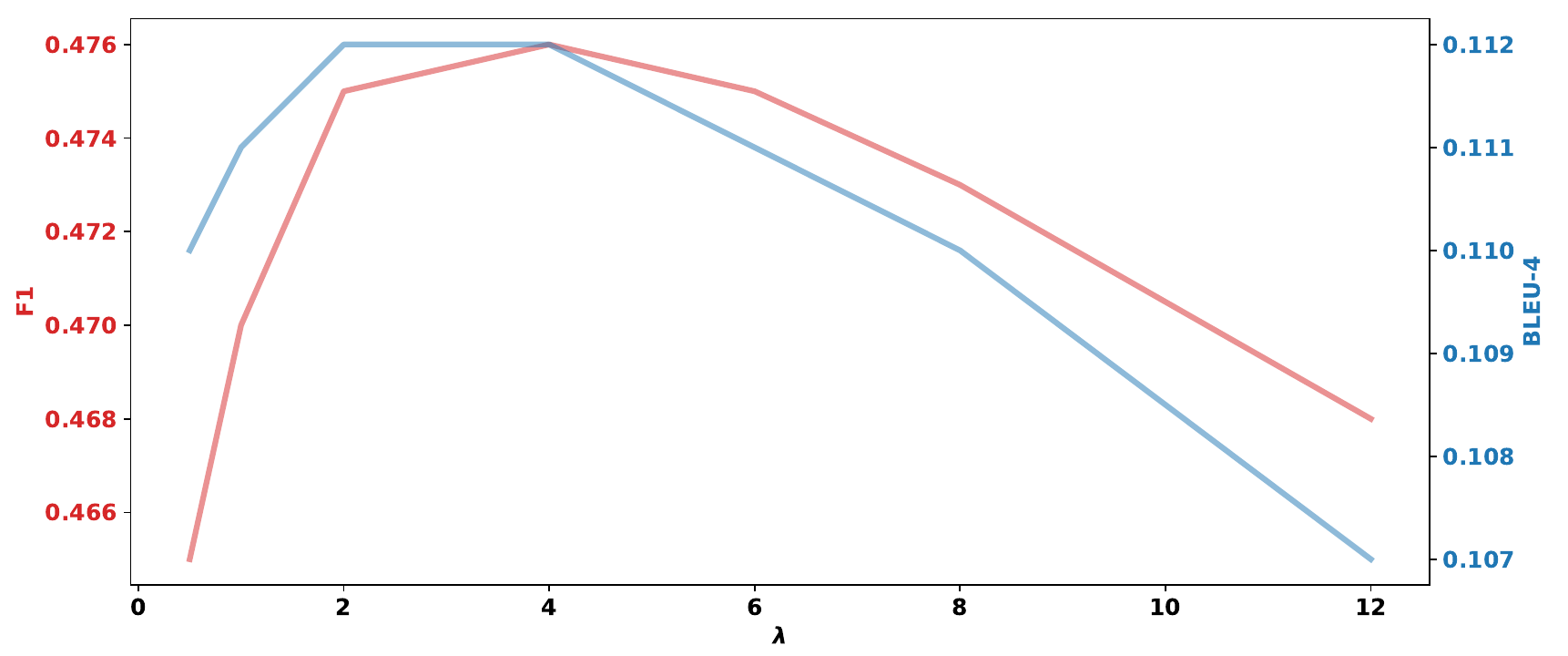}
    \caption{$\lambda$}
  \end{subfigure}
  \hfill
  \begin{subfigure}[b]{1\linewidth}
	\includegraphics[width=\textwidth]{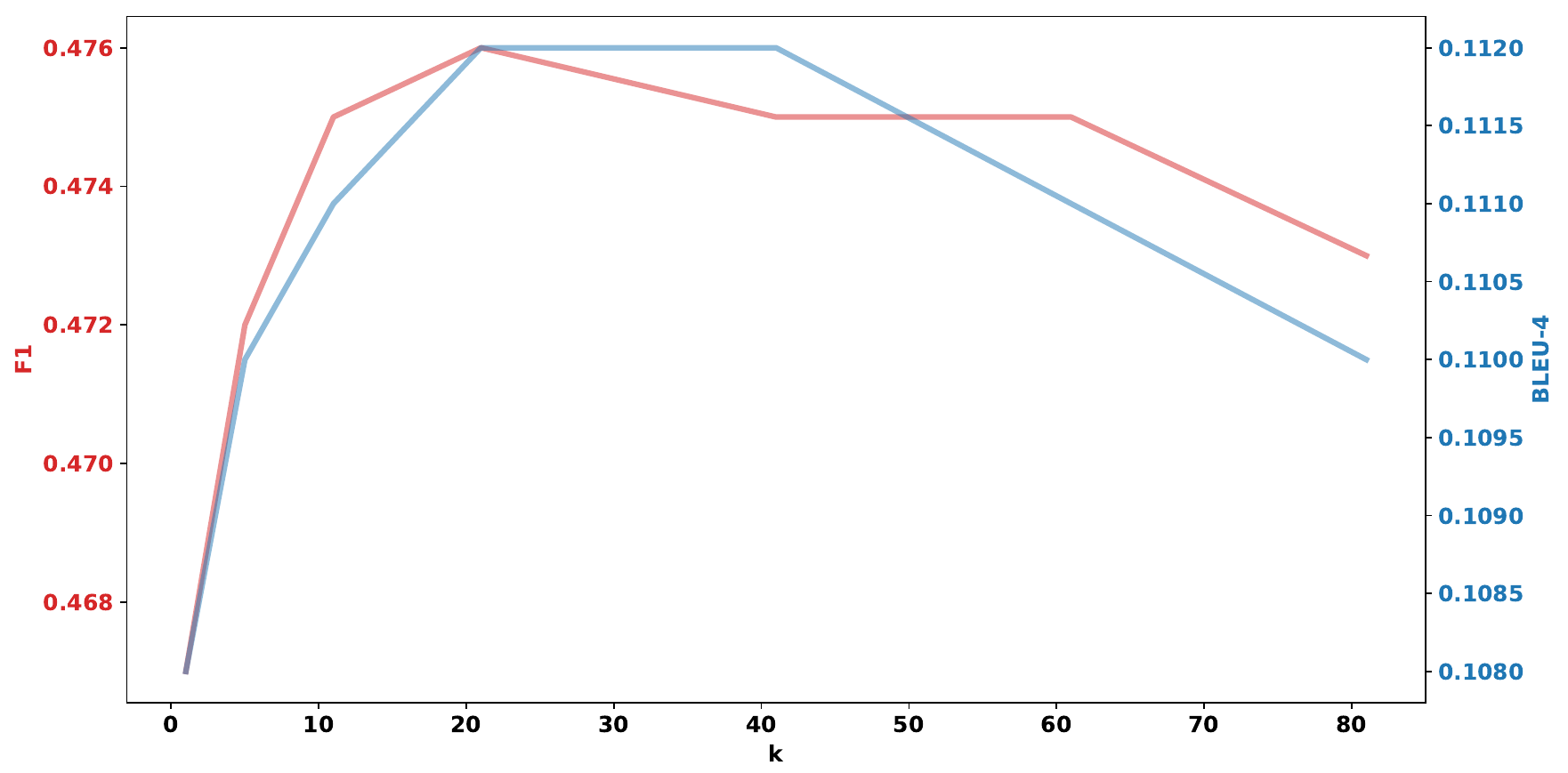}
    \caption{$k$}
  \end{subfigure}
\caption{Analysis of the hyperparameter (a) $\lambda$ and (b) $k$ with respect to F1 and BLEU-4 on the MIMIC test.}
\label{fig:hyper}
\end{figure}

\subsection{N-gram Statistics}
\label{sec8}

To verify whether DDP reduces the generation of high-frequency phrases, we counted the frequency of top-10 high-frequency phrases in the generated reports for method with and without DDP. To obtain these high-frequency phrases, we counted the frequency of 4-grams, 5-grams, 6-grams, and 7-grams in the MIMIC training set, and selected the ten most frequent ones (N-grams with similar meaning were merged and longer N-grams were preferred). Then we counted the frequency of these phrases in the generated reports. Figure 2 and 3 give the statistics on MIMIC test and IU test, respectively. Firstly, we can see from Figure 2 that the method without DDP generates more high-frequency phrases than the one with DDP on all the selected phrases. On average, the number of high-frequency phrases generated by the method without DDP is twice that of the method with DDP. Since the average report length of the two methods are comparable (both around 60), we believe the method with DDP tend to generate more diverse N-grams and it may be due to the extra diagnostic results that provide more information to the text decoder. In contrast, this issue is not prominent on IU test shown in Figure 3. A possible explanation is that the generated IU reports are much shorter than that of MIMIC on average (35 vs. 60), giving the method without DDP less opportunity to generate these high-frequency phrases.   

\begin{table}[t]
  \footnotesize
  \centering
  \newcolumntype{C}{>{\centering\arraybackslash}X}%
  \begin{tabularx}{\linewidth}{llCCC}
    \toprule
    \textbf{Dataset} & \textbf{Model} & \textbf{Precision} & \textbf{Recall}  & \textbf{F1}    \\
    \midrule
    \multirow{7}{*}{\textbf{MIMIC}} & R2Gen$^{\dagger}$ & 0.311 & 0.141  & 0.151  \\ 
    & CVT2Dis.$^{*}$ & 0.256  & 0.382  & 0.307  \\
    & M2KT$^{\dagger}$ & 0.346 & 0.253 & 0.273 \\
    & DCL$^{\dagger}$  & 0.345  & 0.267 & 0.284   \\
    & RGRG$^{*\dagger}$   & 0.355  & 0.340 & 0.318   \\
    \cmidrule(r){2-5}
    & Ours w/o SDL    & \textbf{0.406}  & 0.320 & 0.319    \\
    & Ours w/ SDL   & 0.396  & \textbf{0.393} & \textbf{0.381}    \\
    \midrule
    \multirow{7}{*}{\textbf{IU}} & R2Gen$^{\dagger}$  & 0.161 & 0.104  & 0.071  \\ 
    & CVT2Dis.$^{*\dagger}$ & 0.325  & 0.166  & 0.155  \\
    & M2KT$^{\dagger}$ & 0.224 & 0.179 & 0.151 \\
    & DCL$^{\dagger}$  & 0.257  & 0.202 & 0.177   \\
    & RGRG$^{*\dagger}$   & 0.241  & 0.224 & 0.187   \\
    \cmidrule(r){2-5}
    & Ours w/o SDL    & 0.265  & 0.202 & 0.193    \\
    & Ours w/ SDL   & \textbf{0.275}  & \textbf{0.288} & \textbf{0.246}    \\
    \bottomrule
  \end{tabularx}
  \vspace{-2mm}
  \caption{Comparison to SOTA methods on MIMIC and IU in terms of macro-averaged CE metrics. $*$ indicates the used image size is larger than 224. $\dagger$ indicates the performance evaluated by us. The best results are in \textbf{bold}.}
  \label{tab:results_macro}
\end{table}

\subsection{Macro-Averaged Results}
\label{sec9}

As mentioned in the Disease Balance subsection, macro-averaged CE metrics are better at evaluating balancedness of diseases than example-based CE metrics, because the former average the results over diseases. In this section, we compare with SOTA methods using macro-averaged CE metrics for a disease-balanced CE evaluation. Note that we only compare with SOTA methods that have released their codes, and Table 2 shows the results on both MIMIC and IU test. We can see from the table that the gap between the best existing method and our method is even larger than example-based CE metrics because the disease-balanced problem was explicitly addressed by our method. Specifically, in MIMIC, our absolute improvement over the best existing method RGRG is 6.3\% under the macro-averaged F1 while such an improvement under example-based F1 is 2.9\%. Similarly, in IU, the improvement has been improved from 3.1\% to 5.9\% when the F1 score is switched from example-based to macro-averaged. Additionally, the comparison between our method with and without SDL also verifies the effectiveness of SDL in delivering disease-balanced diagnosis. For example, the absolute improvement on MIMIC F1 has been 6.2\% when SDL is applied, which is a significant advancement.

\begin{figure*}[t]
\centering
  \includegraphics[width=1\linewidth]{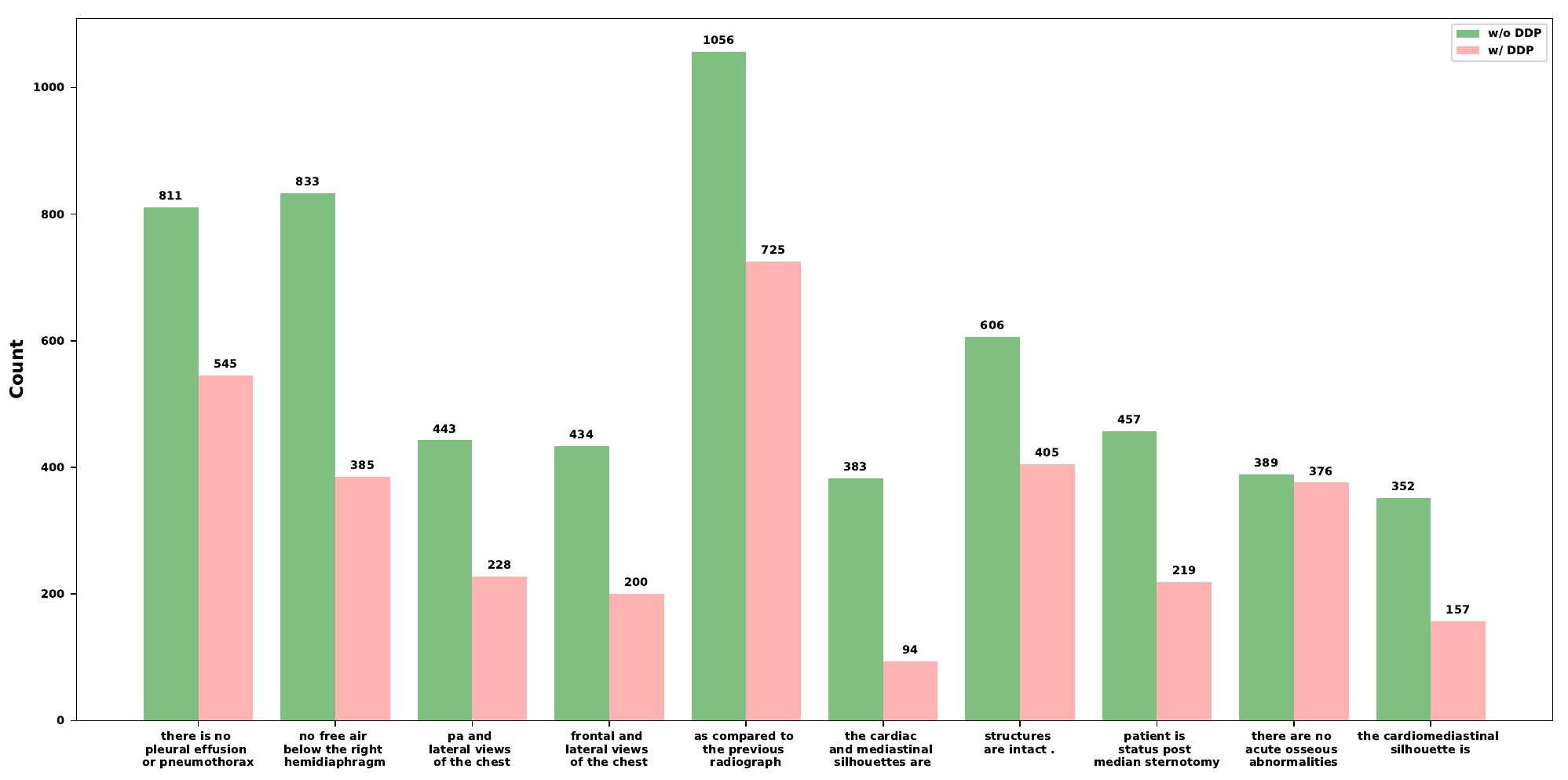}
\vspace{-4mm}
\caption{Comparing the count of high-frequency phrases between the method with and without DDP on the MIMIC test.}
\label{fig:ngram_mimic}      
\end{figure*}

\begin{figure*}[t]
\centering
  \includegraphics[width=1\linewidth]{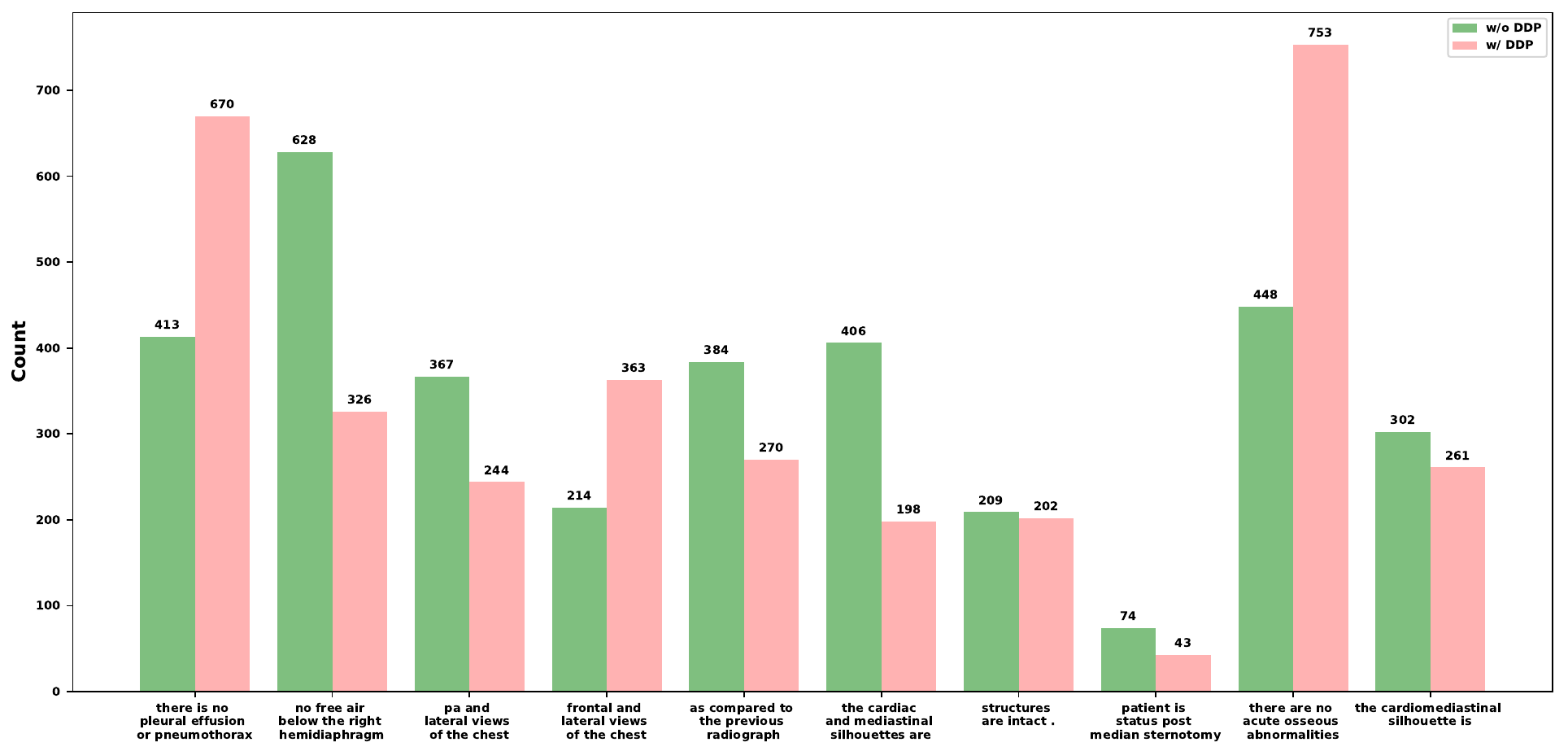}
\vspace{-4mm}
\caption{Comparing the count of high-frequency phrases between the method with and without DDP on the IU test.}
\label{fig:ngram_mimic}      
\end{figure*}

\subsection{More Qualitative Results}
\label{sec10}

We show more qualitative examples in Figure 4. As can be seen, the proposed PromptMRG gives more accurate predictions in disease diagnosis than that of the baseline. For example, the baseline failed to identify the cardiomegaly in the first image while wrongly predicted the enlargement of the cardiac silhouette and pneumonia, which do not exist in the image. In contrast, the predicted diseases of PromptMRG are mostly correct except opacity. Similarly, in the second image, the baseline failed to identify pleural effusion as positive while PromptMRG succeeded. 

\begin{figure*}[t]
\centering
  \includegraphics[width=1\linewidth]{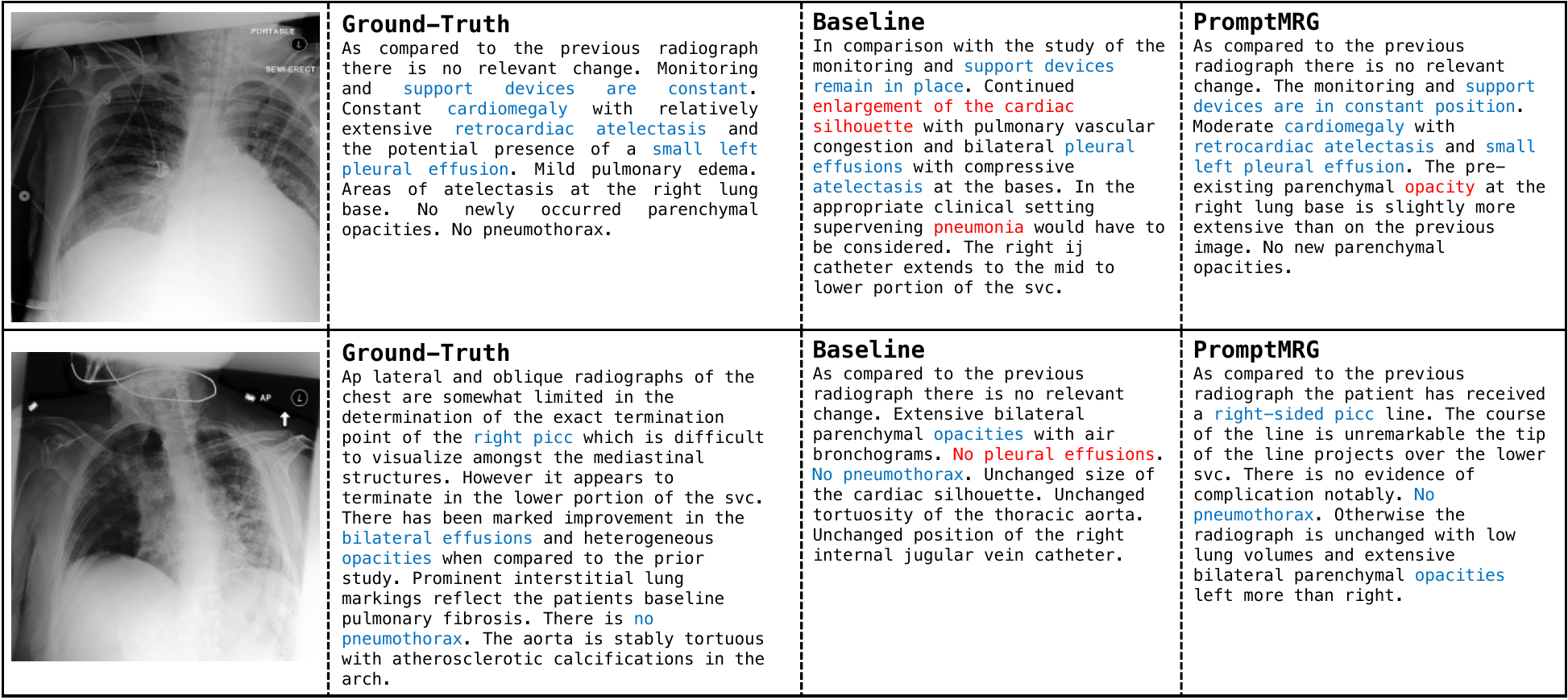}
\vspace{-4mm}
\caption{More qualitative examples of the baseline and the proposed method. Blue font indicates consistent content with the ground-truth while red font indicates incorrect content.}
\label{fig:qual_more}      
\end{figure*}

\end{document}